%% file: main.tex
\documentclass[10pt]{article} 
\usepackage[accepted]{tmlr}

\input{math_commands.tex}

\usepackage{booktabs}
\usepackage{hyperref}
\usepackage{url}
\usepackage{caption}
\usepackage{subcaption}
\usepackage{array}
\usepackage{kotex}

\usepackage{graphicx}

\title{Do Vision Encoders Truly Explain Object Hallucination?: Mitigating Object Hallucination via Simple Fine-Grained CLIPScore}

\author{\name Hongseok Oh \email cxv0519@uos.ac.kr \\
      \addr Department of Artificial intelligence \\
      University of Seoul
      \AND
      \name Wonseok Hwang \email wonseok.hwang@uos.ac.kr \\
      \addr Department of Artificial intelligence \\
      University of Seoul}

\def\month{01}  
\def\year{2026} 
\def\openreview{\url{https://openreview.net/forum?id=JTua6tDPgZ}} 

\begin{document}

\maketitle

\begin{abstract}

Recently, Large Vision-Language Models (LVLMs) show remarkable performance across various domains.
However, these models suffer from object hallucination.
In this work, we study object hallucination primarily in a discriminative, retrieval-style evaluation setting (OHD-Caps), rather than in free-form caption generation.
This study revisits the previous claim that the cause of such hallucinations lies in the limited representational capacity of the vision encoder. 
Our analysis implies that the capacity of the vision encoder is not necessarily a major limiting factor in detecting object hallucination.
Based on this insight, we propose Fine-grained CLIPScore (F-CLIPScore), a simple yet effective evaluation metric that enhances object-level granularity by incorporating text embeddings at the noun level.
Evaluations on the OHD-Caps benchmark show that F-CLIPScore significantly outperforms conventional CLIPScore in accuracy by a large margin of \textbf{39.6\%} without additional training.
We further demonstrate that F-CLIPScore-based data filtering reduces object hallucination in LVLM (4.9\% in POPE accuracy after alignment pretraining).
Our code is publicly available at \url{https://github.com/abzb1/f-clip}

\end{abstract}

\section{Introduction} \label{sec:intro}

Recent studies identify Large Vision-Language Models (LVLMs) as a leading approach for vision-language integration~\citep{LLaVA, qwen2vl, minigpt4, internvl}. However, like hallucinations in large language models (LLMs)~\citep{llm_hallucination, sirenssong}, LVLMs exhibit object hallucination, referring to nonexistent or misidentified objects that may undermine their reliability~\citep{eval_obj_hallucination_lvlm, lvlm_hallucination_survey}.
While object hallucination is often discussed in the context of caption generation, this paper focuses on discriminative vision–language models and retrieval-style evaluation protocols, where the goal is to select or score candidate captions given an image.

\citet{ohd-caps} built OHD-Caps, a dataset designed to measure object hallucination in a discriminative caption-selection setting. The dataset comprises 1.5k image-captions pairs where a model needs to select the best caption that does not show hallucinations. They found that CLIPScore~\citep{clipscore} achieved only 10--20\% accuracy, and the further fine-tuning with the proposed objective function with the dataset improves the accuracy up to 80--90\%. However, when they connected the OHD-Caps-trained CLIP to an LVLM and conducted full fine-tuning, the resulting accuracy sometimes drops, showing lower performance compared to original CLIP (for instance, 1st row of Table 4 in \citep{ohd-caps} shows the accuracy in POPE benchmarks~\citep{pope} drops from 85.4\% to 81.2\%).

\begin{figure}[t]
    \centering
    \includegraphics[width=0.5\linewidth]{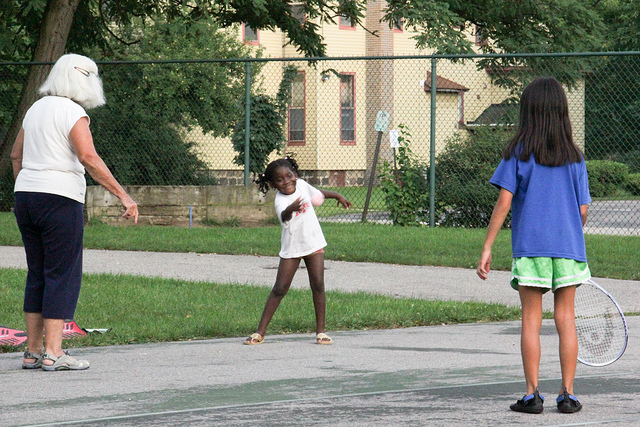}
    \begin{minipage}{0.8\textwidth}
        \small
        \textbf{CLIPScore} (\textbf{w/o training}): \texttt{A lady and two children in the street playing with a tennis racquet, a car nearby, and a chair.} 
        \\\\\textbf{CLIPScore} (\textbf{trained}): \texttt{A lady and two dogs in the park playing with a frisbee.}
       \\\\\textbf{F-CLIPScore} (\textbf{w/o training}): \texttt{A lady and two children in the street playing with a tennis racquet.}
    \end{minipage}
    \caption{A representative example from the OHD-Caps test set is shown. The original CLIP selects a sentence mentioning “children” and “tennis” but adds hallucinated objects. The OHD-Caps-trained CLIP hallucinates “dog” and “frisbee” without introducing new content. In contrast, F-CLIPScore selects a sentence that preserves the original meaning without hallucinations.}
    \label{fig:ohd_sample}
\end{figure}

Moreover, we found that OHD-Caps-trained CLIP sometimes tends to hallucinate by replacing existing objects. When both CLIPScore and OHD-Caps-trained CLIP fail, but our Fine-grained CLIPScore (introduced in a later section) succeeds, a clear pattern emerges: CLIPScore adds nonexistent objects, whereas OHD-Caps trained CLIP replaces existing ones. This trend holds across COCO, Flickr30k, and NoCaps with rates of 56\%, 58\%, and 59\%, respectively. A representative example is shown in Figure~\ref{fig:ohd_sample}. This suggests that object hallucination may stem from factors beyond the vision encoder’s capacity.

To address this issue, we introduce Fine-grained CLIPScore (F-CLIPScore), a novel image-text correlation metric.
F-CLIPScore leverages a sentence parser like spaCy~\cite{spacy} and the forward pass of a Vision-Language Model (VLM) like CLIP~\cite{clip}, offering an efficient way to evaluate the vision encoder’s representational capacity.
Our experimental results show that applying F-CLIPScore to the OHD-Caps test set improves accuracy by \textbf{+39.6\%} without additional training. This indicates that the limited capacity of the vision encoder may not be the primary cause of object hallucination.
Additionally, we verify that using F-CLIPScore for pretraining data curation in LVLMs enables the training of models with reduced hallucination, even with significantly fewer data samples.
Notably, in LVLM pretraining, data filtering alone improved POPE accuracy by 4.9\% compared to the baseline.

This study offers the following key contributions:
\begin{itemize}
    \item We introduce Fine-grained CLIPScore, a novel evaluation metric that relies solely on forward propagation.
    \item We provide evidence suggesting that, in discriminative evaluation, object hallucination is not primarily explained by the vision encoder’s representational capacity alone.
    \item  We demonstrate that F-CLIPScore enables more efficient LVLM training with reduced object hallucination through pretraining data curation.
\end{itemize}

Our code is available at \url{https://github.com/abzb1/f-clip}.

\section{Related Work}

Object hallucination refers to cases in which the generated textual descriptions include objects that do not correspond to the given image~\citep{lvlm_hallucination_survey}.
LVLMs generally consist of three components: a vision encoder, an LLM, and an adapter~\citep{LLaVA}.
The structural characteristics of LVLMs contribute to object hallucination, which arises from multiple intertwined factors~\citep{lvlm_hallucination_survey}.
While some studies argue that hallucinations can be mitigated by enhancing the decoding process of the LLM~\citep{lcd_lvlm, icd_lvlm}, others suggest that one of the causes lies in the limited representational capacity of the vision encoder~\citep{ohd-caps}.
Additionally, some research indicates that training the adapter with contrastive data is essential to reduce object hallucination~\citep{hacl}.
Moreover, the trained bias of the model has also been identified as a cause of hallucination~\citep{ciem, robust_instruction_tuning}.

CLIPScore~\citep{clipscore} is a reference-free evaluation metric that assesses the consistency between an image and text caption by computing the cosine similarity between the embeddings generated by the vision encoder and text encoder of the CLIP model. Beyond its application in measuring caption quality, several studies have also leveraged CLIPScore for data curation in the training of Vision-Language Models (VLMs)~\citep{laion, datacomp}.

A recent study utilized CLIPScore to evaluate object hallucination in Vision-Language Models (VLMs)~\citep{ohd-caps}. Their findings suggest that this phenomenon stems from the limited capacity of the vision encoder.
In this study, we carefully reassess this claim and demonstrate that object hallucination is not necessarily caused by the limitations of the vision encoder alone.

\section{Methods}
\subsection{Motivation}

\begin{figure}[t]
    \centering
    \includegraphics[width=0.9\linewidth]{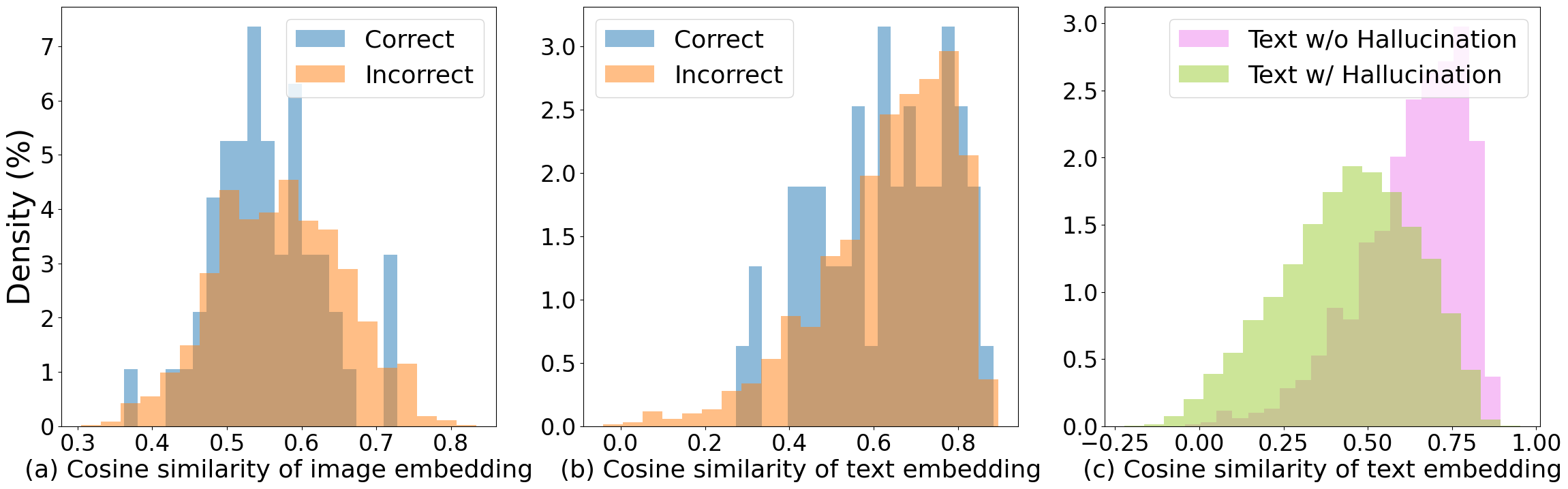}
    \caption{
    Histograms of cosine similarity between two embedding vectors: one from the original CLIP-L and the other from the OHD-Caps-trained CLIP-L.
    (a) The histogram from the vision encoders. \texttt{Correct} (blue) indicates the scores are from the examples where original CLIP-L predict the ground truth. The other examples are colored in orange.
    (b) The histogram from the text encoders. Same color scheme is employed. Measured only on ground truth text.
    (c) The cosine similarity distribution between text embeddings of text without object hallucination (purple) and with object hallucination text (green) for all samples.
    }
    \label{fig:cos_sims}
    \vspace{-5mm}
\end{figure}

Based on our initial observation that OHD-Caps-trained CLIP occasionally does not yield better results (Section \ref{sec:intro}), we further investigate how fine-tuning affects the embedding vectors produced by vision and text encoders of CLIP.
We compute the cosine similarity between two embedding vectors: one from the original CLIP-L and the other from the OHD-Caps-trained CLIP-L, using image-text pairs in the OHD-Caps test set.

We first computed the distribution of cosine similarity between the image embeddings from CLIP-L and OHD-Caps-trained CLIP-L, focusing on the samples that were correctly predicted by CLIP-L. We then performed the same analysis on the samples that were incorrectly predicted by CLIP-L and compared the two distributions.
As shown in Figure~\ref{fig:cos_sims}a, we observe that there is no substantial difference between the two distributions.
We conducted the same analysis using ground-truth text embeddings of the pre-trained and fine-tuned models, and similarly found no significant difference in the cosine similarity distributions between the samples that were correctly and incorrectly predicted by CLIP-L (Figure~\ref{fig:cos_sims}b).
Two-sample t-tests yield p-values of $0.11$ for a and $0.67$ for b.
In contrast, Figure~\ref{fig:cos_sims}c reveals significant changes in text embeddings for captions with hallucination (green), and without hallucination (purple) highlighting the distinct adaptation of text representation.
Two-sample t-tests yield near zero p-value for c.
These observations suggest that OHD-Caps training induces more noticeable changes in the text representation space, particularly in discriminating hallucinated from non-hallucinated captions, whereas the vision representation space exhibits relatively minor changes.

\subsection{Fine-grained CLIPScore}

\begin{figure}
    \centering
    \includegraphics[width=0.8\linewidth]{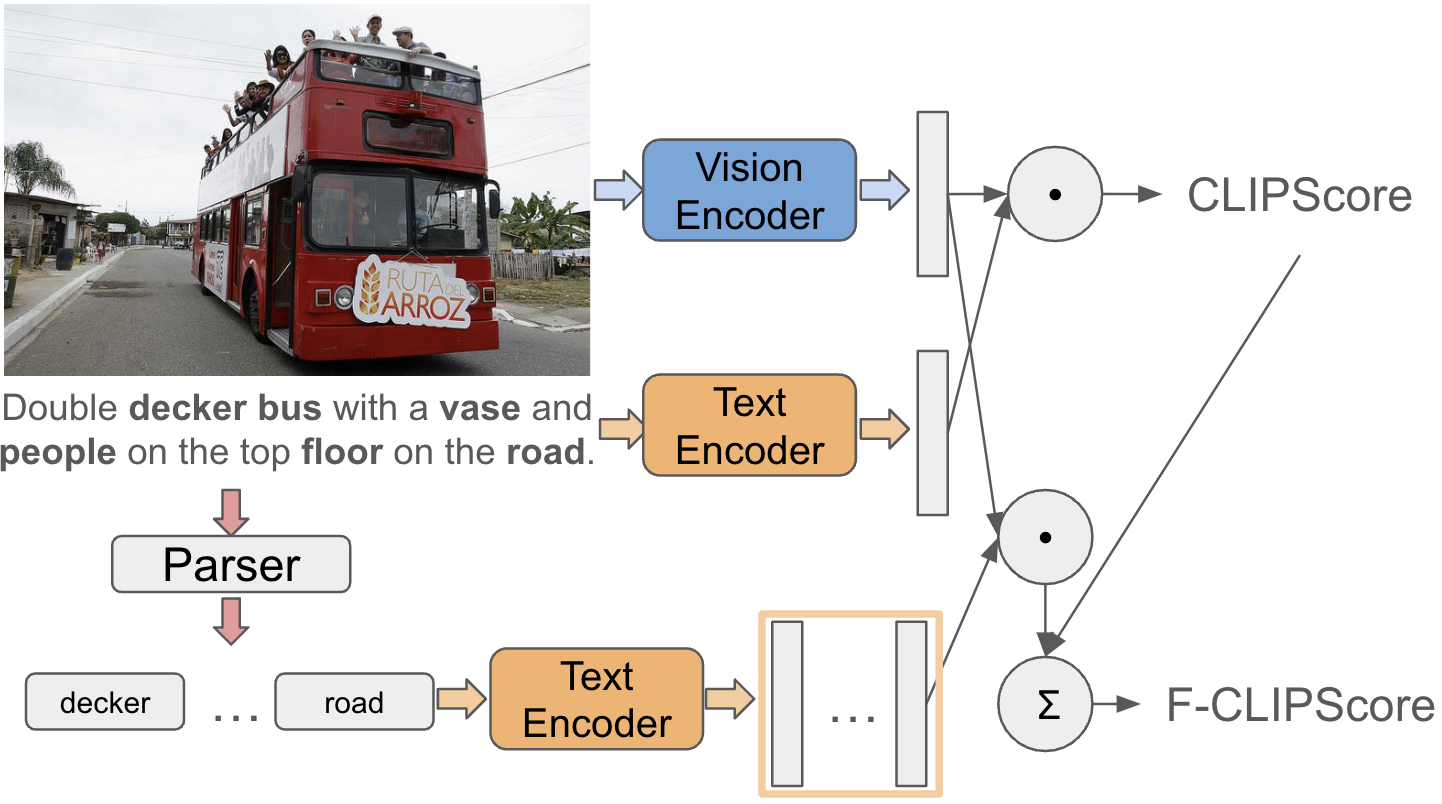}
    \caption{The graphical representation of F-CLIPScore.}
    \label{fig:schema}
\end{figure}

Motivated by these observations, we propose a simple metric called Fine-grained CLIPScore (F-CLIPScore), which enhances the discriminative power of the VLM in order to utilize textual information with more granularity without additional training.
F-CLIPScore first utilizes the spaCy parser~\cite{spacy} to extract nouns from a given sentence.
Then, it evaluates the quality of an image caption by averaging the CLIPScore of the entire sentence and each individual noun (Figure \ref{fig:schema}).

Mathematically, given an image $I$ and a caption $C$ containing a total of $N$ nouns, each denoted as $n_i$, F-CLIPScore is defined as Eq.~\ref{eq:fclip}.

\begin{equation}
    \small
    \operatorname{F-CLIPScore}(I, C) = \frac{\operatorname{CLIPScore}(I, C) + \sum_{i=1}^{N}{\operatorname{CLIPScore}(I, n_i)}}{N+1}
    \label{eq:fclip}
\end{equation}

\section{Experiments}

For evaluating OHD-Caps test set, F-CLIPScore only requires an additional parsing step (we used en\_core\_web\_sm  of spaCy~\citep{spacy}), which takes on average 170 ms per sentence and can be parallelized across samples. We note that this parser runs in linear time with respect to the length of the sentence~\citep{spacyparser}.
For training CLIP with Eq.~\ref{eq:f-cliploss} on the OHD-Caps train set, we used a batch size of 64 and a learning rate of 1e-5, which required 3 hours on an H100 GPU.  
For LLaVA pretraining, we employed an effective batch size of 256 and a learning rate of 1e-3, which took 7 hours on two H100 GPUs.

\section{Results}

By utilizing the F-CLIPScore, which directly leverages the image embeddings from the CLIP vision encoder without any training or gradients, while only adding a parsing step during forward propagation, we can efficiently gain insights into whether the issue of object hallucination arises from the limited capability of the vision encoder.

\subsection{F-CLIPScore on OHD-Caps} \label{subsec:ohdcaps}

We evaluated the OHD-Caps test set, an object hallucination assessment dataset, using the proposed F-CLIPScore.
As shown in Table~\ref{table:ohd_caps_test}, evaluation results with OpenAI CLIP ViT-L~\cite{clip} indicate that F-CLIPScore outperformed the baseline model by up to 39.6\% without additional training (row 4 vs. 5). However, it still performed 17.6\% to 37.4\% worse than trained models (row 5 vs. 7). This may indicate that although CLIP vision encoders may not be the main cause of object hallucination, further training may enhance their capability.

\begin{table}[tb]
    \centering
    \resizebox{0.5\columnwidth}{!}{
    \begin{tabular}{l|c|c|c}
        \toprule
        \multicolumn{4}{c}{OHD-Caps ACC (\%, $\uparrow$)} \\
        \midrule
        \textbf{Metric} & \textbf{COCO} & \textbf{Flickr30k} & \textbf{NoCaps} \\
        \midrule
        \multicolumn{4}{c}{w/o training} \\
        \midrule
        CLIPScore & 22.6 & 22.6 & 12.4 \\
        $\operatorname{F-CLIPScore}$ & \textbf{62.2} & \textbf{62.2} & \textbf{46.6} \\
        \hline
        \multicolumn{4}{c}{trained w/ OHD-loss$^\dagger$} \\
        \midrule
        CLIPScore & 79.8 $_{\pm 1.7}$ & \textbf{84.8} $_{\pm 1.6}$ & 84.0 $_{\pm 0.7}$ \\
        \midrule
        \multicolumn{4}{c}{trained w/ F-CLIPScore loss} \\
        \midrule
        CLIPScore & \textbf{80.5} $_{\pm 1.8}$ & \textbf{84.8} $_{\pm 1.3}$ & \textbf{84.1} $_{\pm 1.3}$ \\
        \bottomrule
    \end{tabular}
    }\\
    \small
    $\dagger$: We trained openai/clip-vit-large-patch14 with the train code from~\cite{ohd-caps}
    \caption{Accuracy on the OHD-Caps test set evaluated with OpenAI CLIP-L. For trained models, we report the average evaluation results over 10 runs with different random seeds.}
    \label{table:ohd_caps_test}
\end{table}

To test whether F-CLIPScore is orthogonal to the previously proposed method~\cite{ohd-caps}, we modify the loss as
\begin{equation}
    \small
    L_{CLIP}+L_{OHD}+\frac{\alpha}{B}\sum_{i=1}^{B}{(1-\operatorname{F-CLIPScore}(I_i, C_i))}
    \label{eq:f-cliploss}
\end{equation}
where $L_{CLIP}$ is the contrastive loss proposed in ~\cite{clip}, and $L_{OHD}$ is the marginal loss proposed in ~\cite{ohd-caps}. $B$ denotes the batch size, while $I_i$ and $C_i$ are image and caption of the $i$-th positive pair. $\alpha$ is a hyperparameter that we set to $0.3$.
Applying F-CLIPScore as a loss improved performance by 0.7 percentage points on COCO and 0.1 percentage points on NoCaps (row 7 vs. 9).
These results suggest that F-CLIPScore could be a complementary component in VLM training.

We randomly replaced the caption embeddings into noun embeddings extracted from a Wikipedia corpus~\citep{davies2015wikipedia} to examine how F-CLIPScore responds to such perturbations. As shown in Table~\ref{tab:fclip_ablation}, we observed the expected decline in accuracy as nouns were replaced. This indicates that the strong performance of F-CLIPScore, even without additional training, is not due to random noise, but rather stems from its ability to more effectively leverage fine-grained textual embeddings.

\begin{table}[tb]
    \centering
    \small
    \resizebox{0.5\columnwidth}{!}{
        \begin{tabular}{ccccc}
            \toprule
            \multicolumn{5}{c}{\textbf{Replacement Rate}} \\
            \midrule
            0.2 & 0.4 & 0.6 & 0.8 & 1.0 \\
            \midrule
            \multicolumn{5}{c}{\textbf{COCO Acc mean $\pm$ std (\%, $\uparrow$)}} \\
            \midrule
             38.9$\pm$2.5 & 35.9$\pm$1.4 & 31.0$\pm$1.7 & 23.5$\pm$2.0 & 19.9$\pm$1.3 \\
            \midrule
            \multicolumn{5}{c}{\textbf{Flickr30k Acc mean $\pm$ std (\%, $\uparrow$)}} \\
            \midrule
            37.5$\pm$1.5 & 33.0$\pm$1.2 & 26.2$\pm$2.1 & 22.0$\pm$1.6 & 18.9$\pm$3.0 \\
            \midrule
            \multicolumn{5}{c}{\textbf{NoCaps Acc mean $\pm$ std (\%, $\uparrow$)}} \\
            \midrule
            29.4$\pm$1.9 & 24.2$\pm$2.3 & 21.2$\pm$2.0 & 16.2$\pm$1.7 & 16.5$\pm$0.4 \\
            \bottomrule
        \end{tabular}
        }
    \caption{F-CLIPScore results on the OHD-Caps test set with different random noun replacement rates (0.2 to 1.0). Each value shows the mean accuracy $\pm$ standard deviation over five random seeds.}
    \label{tab:fclip_ablation}
\end{table}

\begin{table}[tb]
    \centering
    \small
    \resizebox{0.5\columnwidth}{!}{
        \begin{tabular}{l|c|c|c}
            \toprule
            \multicolumn{4}{c}{OHD-Caps ACC (\%, $\uparrow$)} \\
            \midrule
            \textbf{Metric} & \textbf{coco} & \textbf{flickr30k} & \textbf{nocaps} \\
            \midrule
            CLIPScore & 22.6 & 22.6 & 12.4 \\
            $\operatorname{F-CLIPScore}_{V}$ & 23.6 & 24.8 & 15.8 \\
            $\operatorname{F-CLIPScore_{NP}}$ & 54.2 & 57.8 & 39.6 \\
            $\operatorname{F-CLIPScore}_{N}$ & \textbf{62.2} & \textbf{62.2} & \textbf{46.6} \\
            \bottomrule
        \end{tabular}
    }
    \caption{Accuracy on the OHD-Caps test set evaluated with OpenAI CLIP-L. $\operatorname{F-CLIPScore}_{N}$ refers to the noun-based $\operatorname{F-CLIPScore}$ defined in Eq.~\ref{eq:fclip}. $\operatorname{F-CLIPScore_{NP}}$ uses noun phrases instead of individual nouns, and $\operatorname{F-CLIPScore}_{V}$ replaces the nouns with verbs.}
    \label{table:ohd_caps_test2}
\end{table}

We experimented with alternative configurations of F-CLIPScore using verbs and noun phrases instead of nouns. As shown in Table~\ref{table:ohd_caps_test2}, using verbs yielded little to no improvement, while noun phrases performed better than verbs but still fell short of the performance achieved with nouns.

\begin{table}[tb]
    \centering
    \resizebox{0.65\columnwidth}{!}{
    \begin{tabular}{lccc}
        \hline
        \textbf{OHD-Caps Accuracy} & \textbf{COCO} & \textbf{Flickr30k} & \textbf{NoCaps} \\
        \hline
        CLIP-B/32 (CLIPScore) & 15.2 & 17.6 & 10.2 \\
        CECLIP \citep{ceclip}               & 32.8 & 28.0 & 25.0 \\
        NegCLIP \citep{negclip}              & 52.8 & 40.8 & 23.4 \\
        \textbf{CLIP-B/32 (F-CLIPScore)} & \textbf{56.0} & \textbf{53.2} & \textbf{43.0} \\
        \hline
    \end{tabular}
    }
    \caption{OHD-Caps accuracy across different methods.}
    \label{table:ohd_caps_different_methods}
\end{table}

Our F-CLIPScore even outperforms the training-based methods \citep{ceclip, negclip}, as shown in Table~\ref{table:ohd_caps_different_methods}.

\begin{table}[tb]
    \centering
    \resizebox{0.7\columnwidth}{!}{
        \begin{tabular}{l|ccc}
            \toprule
            \textbf{OHD-Caps Accuracy} & \textbf{COCO} & \textbf{Flickr30k} & \textbf{NoCaps} \\
            \midrule
            SigLIP ViT-L (CLIPScore) \citep{siglip} & 48.6 & 38.4 & 30.2 \\
            \textbf{SigLIP ViT-L (F-CLIPScore)} & \textbf{69.0} & \textbf{64.2} & \textbf{54.2} \\
            EVA-CLIP ViT-L (CLIPScore) \citep{eva-clip} & 38.6 & 31.8 & 22.6 \\
            \textbf{EVA-CLIP ViT-L (F-CLIPScore)} & \textbf{69.6} & \textbf{64.8} & \textbf{55.4} \\
            \midrule
            CLIP ViT-B (CLIPScore)     & 15.2 & 17.6 & 10.2 \\
            \textbf{CLIP ViT-B (F-CLIPScore)} & \textbf{56.0} & \textbf{53.2} & \textbf{43.0} \\
            CLIP ViT-L (CLIPScore)     & 22.6 & 22.6 & 12.4 \\
            \textbf{CLIP ViT-L (F-CLIPScore)} & \textbf{62.2} & \textbf{62.2} & \textbf{46.6} \\
            CLIP ViT-H (CLIPScore)     & 36.8 & 31.4 & 20.6 \\
            \textbf{CLIP ViT-H (F-CLIPScore)} & \textbf{57.6} & \textbf{59.4} & \textbf{46.6} \\
            \bottomrule
        \end{tabular}
    }
    \caption{OHD-Caps accuracy across different vision-encoder architectures and sizes.}
    \label{table:ohd_caps_across_scale_and_model}
\end{table}

Furthermore, Table~\ref{table:ohd_caps_across_scale_and_model} demonstrates that F-CLIPScore consistently achieves gains across different scales and architectures of vision-language models.

\subsection{LVLM Pretrain Data Curation with F-CLIPScore}

As shown in Section~\ref{subsec:ohdcaps}, F-CLIPScore was able to exhibit competent performance in detecting object hallucination without any training, and could further serve as an complementary method for training VLMs.
We aimed to investigate whether F-CLIPScore can influence the pretraining process of LVLM by acting as a data filter between the vision encoder and the LLM. 
To this end, we applied F-CLIPScore to filter the pretraining data for LLaVA~\citep{LLaVA}, which consists of 558k samples, by removing the bottom $x\%$ of the alignment training set based on F-CLIPScore.

\begin{table}[tb]
    \centering
    \resizebox{0.6\columnwidth}{!}{
    \begin{tabular}{l|ccccc}
        \toprule
        \textbf{POPE Acc ($\uparrow$, \%)} & \multicolumn{5}{|c}{\textbf{Filtering rate (\%)}} \\
        \midrule
        \textbf{Filtering Method} & \textbf{20} & \textbf{30} & \textbf{40} & \textbf{50} & \textbf{60} \\
        \midrule
        Random & 50.7 & 49.9 & 49.9 & 50.5 & 49.8 \\
        \midrule
        CLIPScore (w/o train) & 52.2 & 47.3 & 50.0 & 53.1 & 50.0 \\
        F-CLIPScore (w/o train)  & 51.6 & \textbf{55.5} & 50.6 & 50.1 & 51.8 \\
        CLIPScore (trained) & 49.8 & 49.8 & 50.8 & 49.8 & 52.6 \\
        \midrule
        w/o filtering & \multicolumn{5}{|c}{50.6} \\
        \bottomrule
    \end{tabular}
    }
    \caption{Accuracy on the POPE benchmark after LLaVA pretraining (vision-text alignment training before SFT). We use CLIP-L~\cite{clip} as a vision encoder, and Llama 2 7B~\cite{llama2} as an LLM backbone. ``trained'' indicates the OHD-Caps-trained CLIP-L. ``Random'' refers to the removal of $x\%$ of samples chosen at random.}
    \label{table:pope}
\end{table}

As shown in Table~\ref{table:pope}, training the alignment model on the top 70\% of data curated by F-CLIPScore resulted in a +4.9\% improvement in POPE accuracy compared to training on the entire dataset (row 5 vs. 7).  
In contrast, using the OHD-Caps-trained CLIP for filtering yielded only marginal gains, and random filtering showed no improvement.  
These results may suggest that F-CLIPScore effectively captures object hallucination-related quality, even in general-purpose datasets.  
This finding underscores the need to explore alternative causes of object hallucination beyond the capacity of the vision encoder.
Although model performance shows a non-linear trend with respect to the filtering ratio, our experiments used a fixed number of training epochs, which may account for this behavior. As \citet{Goyal_2024_CVPR} highlights the need to balance data quality with computational cost, identifying the optimal filtering ratio under such trade-offs remains an important direction for future research.

\section{Conclusion}

We introduce F-CLIPScore, a simple yet effective metric for evaluating fine-grained image-caption alignment and addressing object hallucination in Vision-Language Models.
Unlike conventional CLIPScore, which relies solely on sentence-level embeddings, F-CLIPScore also incorporates noun-level embeddings.
This refinement allows the model to better mitigate object hallucination without requiring additional training for the vision encoder. 
We validate F-CLIPScore by showing a +39.6\% accuracy in OHD-Caps benchmark. We also show that data filtering based on F-CLIPScore can enhance LVLM performance in hallucination mitigation, even with a reduced dataset.
Our results suggest that the limitations of existing evaluation metrics, rather than the vision encoder itself, may contribute to object hallucination because they fail to accurately reflect the vision encoder’s true capacity.

\section*{Limitations}

First, our study primarily targets discriminative evaluation and retrieval-style scoring (e.g., OHD-Caps, POPE) and does not directly evaluate hallucination in free-form caption generation. Investigating how noun-level scoring can be integrated into generative decoding or training objectives remains future work.

While this study proposes a method for analyzing and mitigating object hallucination using F-CLIPScore, it is subject to the following limitations. First, we were unable to conduct experiments on the Supervised Fine-Tuning (SFT) for Large Vision-Language Models (LVLMs). In the LVLM training pipeline, after the alignment pretraining phase—where the vision encoder and LLM remain frozen—the SFT stage follows, in which these components are unfrozen and further trained. However, due to computational resource limitations, we did not fully explore the potential impact of F-CLIPScore during SFT. Future work could investigate ways to incorporate F-CLIPScore into the SFT process to enhance training effectiveness.

Second, our method faces linguistic constraints and challenges in multilingual generalization. This study employs the spaCy parser~\citep{spacy} to extract nouns from text, a technique that performs relatively reliably in well-structured languages such as English. However, parsing accuracy may vary across different languages, potentially leading to inconsistencies in F-CLIPScore computation. To address this, future research should explore the scalability of F-CLIPScore by evaluating its effectiveness on multilingual datasets and refining the parsing methodology for broader linguistic applicability.

\subsubsection*{Acknowledgments}
This work was supported by the National Research Foundation of Korea (NRF) grant funded by the Korea government(MSIT) (RS-2025-23524855).

\bibliography{main}
\bibliographystyle{tmlr}

\appendix

\newpage

\section{Qualitative Analysis of F-CLIPScore Filtering}

\begin{figure}[tbh]
  \centering

  \begin{minipage}{0.8\linewidth}
    \centering

    \begin{subfigure}[t]{0.17\linewidth}
      \centering
      \includegraphics[width=\linewidth]{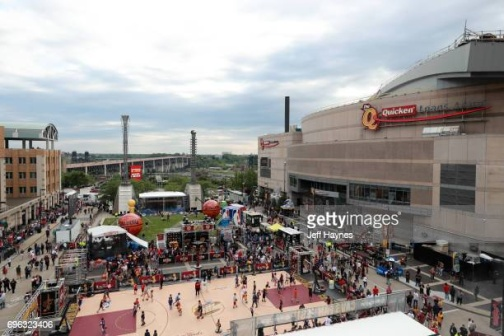}
      \caption{\scriptsize a view of the crowd during the 2016 nba allstar game at madison arena on august 30 2018 in new}
    \end{subfigure}\hfill
    \begin{subfigure}[t]{0.17\linewidth}
      \centering
      \includegraphics[width=\linewidth]{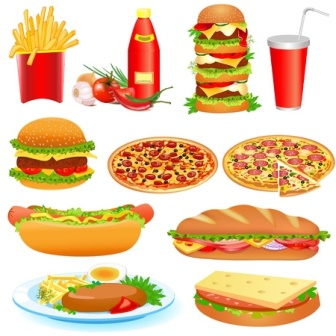}
      \caption{\scriptsize a group of food vectors on a white background}
    \end{subfigure}\hfill
    \begin{subfigure}[t]{0.17\linewidth}
      \centering
      \includegraphics[width=\linewidth]{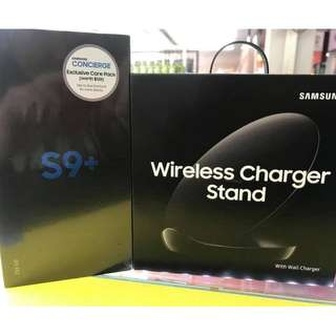}
      \caption{\scriptsize a pair of samsung charger stand with charging and charger in the background}
    \end{subfigure}\hfill
    \begin{subfigure}[t]{0.17\linewidth}
      \centering
      \includegraphics[width=\linewidth]{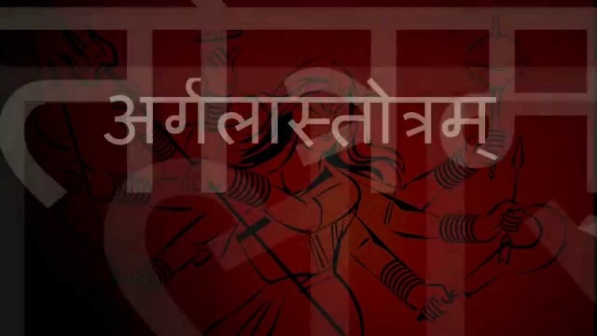}
      \caption{\scriptsize an illustration of lord krishna in marathi language}
    \end{subfigure}\hfill
    \begin{subfigure}[t]{0.17\linewidth}
      \centering
      \includegraphics[width=\linewidth]{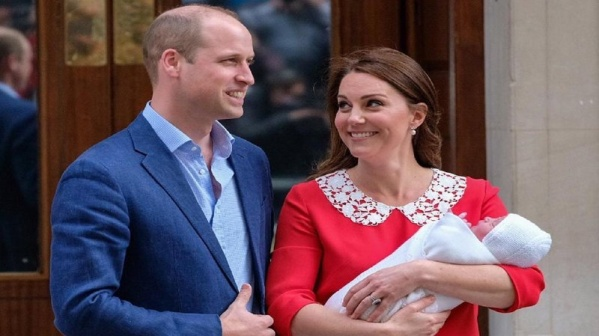}
      \caption{\scriptsize the duke and duchess of cambridge pose for a family portrait}
    \end{subfigure}

    \begin{subfigure}[t]{0.17\linewidth}
      \centering
      \includegraphics[width=\linewidth]{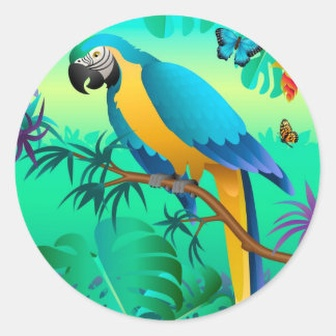}
      \caption{\scriptsize blue and yellow macaw in the jungle sticker}
    \end{subfigure}\hfill
    \begin{subfigure}[t]{0.17\linewidth}
      \centering
      \includegraphics[width=\linewidth]{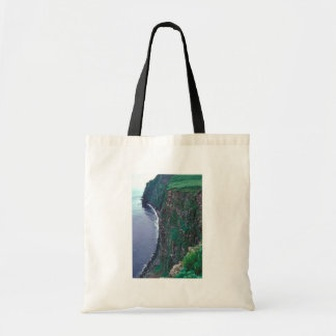}
      \caption{\scriptsize a cliff cliff is on the island of maui canvas bag}
    \end{subfigure}\hfill
    \begin{subfigure}[t]{0.17\linewidth}
      \centering
      \includegraphics[width=\linewidth]{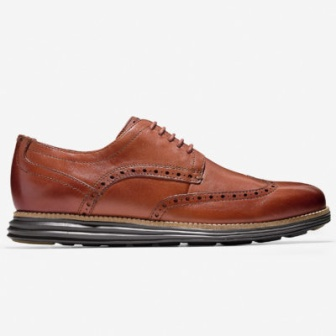}
      \caption{\scriptsize the brown shoe is shown with two black soles and a black rubber outsole}
    \end{subfigure}\hfill
    \begin{subfigure}[t]{0.17\linewidth}
      \centering
      \includegraphics[width=\linewidth]{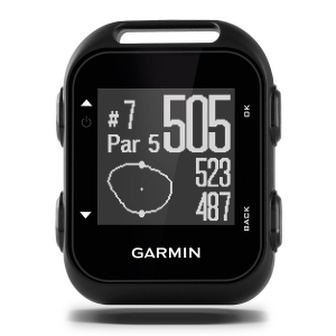}
      \caption{\scriptsize garmin edge 20 gps watch with heart rate and gps coordinates}
    \end{subfigure}\hfill
    \begin{subfigure}[t]{0.19\linewidth}
      \centering
      \includegraphics[width=\linewidth]{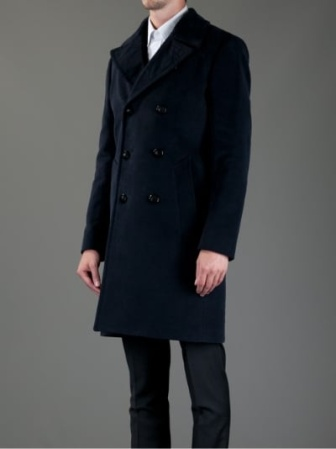}
      \caption{\scriptsize ayrmoors - long length coat 8}
    \end{subfigure}

    \vspace{0.5em}(I) Image samples filtered by CLIPScore at the 30\% filtering rate\par\vspace{0.5em}
    
  \end{minipage}

  \begin{minipage}{0.8\linewidth}
    \centering

    \begin{subfigure}[t]{0.17\linewidth}
      \centering
      \includegraphics[width=\linewidth]{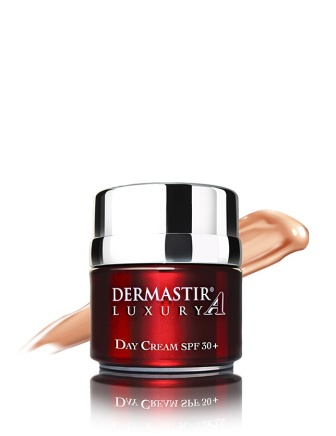}
      \caption{\scriptsize dermastir'd® lux ultra k wax cream spf}
    \end{subfigure}\hfill
    \begin{subfigure}[t]{0.17\linewidth}
      \centering
      \includegraphics[width=\linewidth]{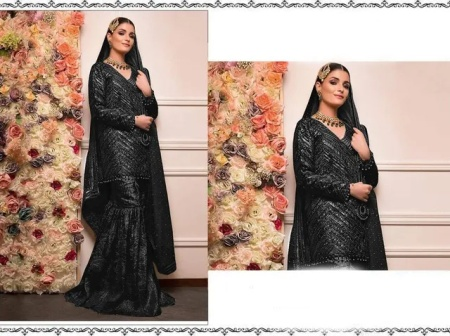}
      \caption{\scriptsize an elegant black designer designer gown with beautiful lace work}
    \end{subfigure}\hfill
    \begin{subfigure}[t]{0.17\linewidth}
      \centering
      \includegraphics[width=\linewidth]{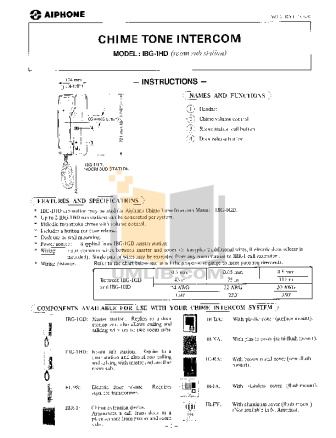}
      \caption{\scriptsize pdf manual for microphone microphones for alto telephone}
    \end{subfigure}\hfill
    \begin{subfigure}[t]{0.17\linewidth}
      \centering
      \includegraphics[width=\linewidth]{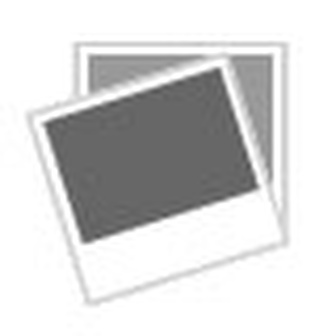}
      \caption{\scriptsize a stainless steel sink with three drainers and two spout faucets}
    \end{subfigure}\hfill
    \begin{subfigure}[t]{0.17\linewidth}
      \centering
      \includegraphics[width=\linewidth]{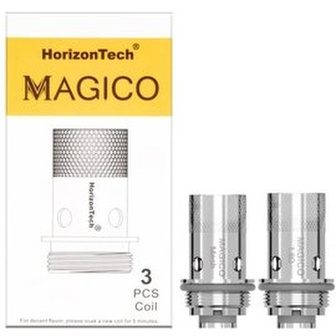}
      \caption{\scriptsize horizon™ magico mesh coil}
    \end{subfigure}

    \begin{subfigure}[t]{0.17\linewidth}
      \centering
      \includegraphics[width=\linewidth]{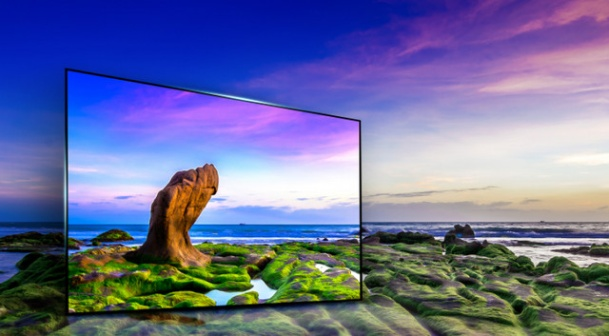}
      \caption{\scriptsize tv on the stone beach on sunset stock photo}
    \end{subfigure}\hfill
    \begin{subfigure}[t]{0.17\linewidth}
      \centering
      \includegraphics[width=\linewidth]{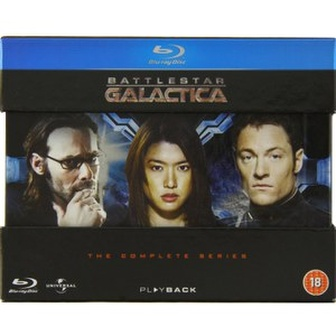}
      \caption{\scriptsize battlestard galactica the complete series blu - ray}
    \end{subfigure}\hfill
    \begin{subfigure}[t]{0.17\linewidth}
      \centering
      \includegraphics[width=\linewidth]{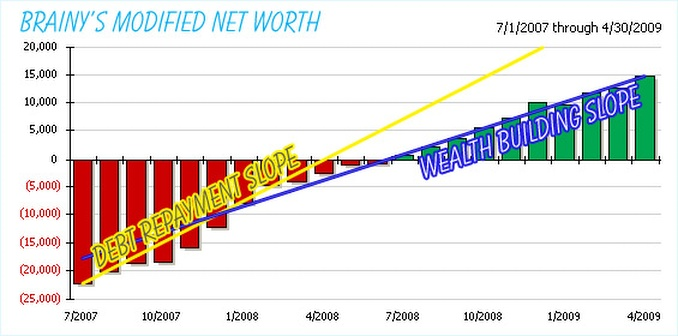}
      \caption{\scriptsize an upward chart shows the number of complaints about brain disease}
    \end{subfigure}\hfill
    \begin{subfigure}[t]{0.17\linewidth}
      \centering
      \includegraphics[width=\linewidth]{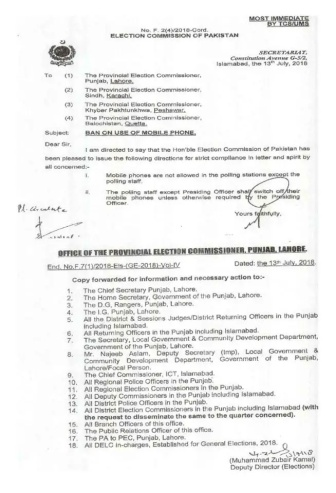}
      \caption{\scriptsize the official documents of the commission for commissioning and remediarnation of power transmission units.png}
    \end{subfigure}\hfill
    \begin{subfigure}[t]{0.17\linewidth}
      \centering
      \includegraphics[width=\linewidth]{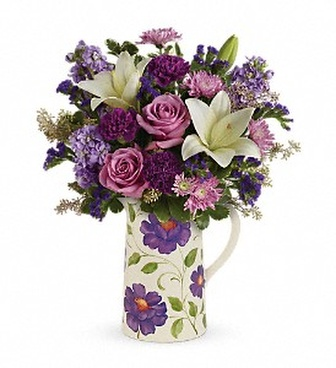}
      \caption{\scriptsize telefish garden pitcher in new castle pa, the bouquet}
    \end{subfigure}

    \vspace{0.5em}(II) Image samples filtered by F-CLIPScore at the 30\% filtering rate\par

  \end{minipage}

  \caption{Randomly sampled 10 image-caption pairs each from (I) samples filtered by CLIPScore and (II) samples filtered by F-CLIPScore at the 30\% filtering rate. Images overlapping between the two metrics were excluded.}
  \label{fig:30_images}
\end{figure}

We first observed the degree of overlap between the F-CLIPScore and CLIPScore metrics when filtering the data.  
At filtering rates of 20, 30, 40, 50, and 60, the proportions of overlapping image–caption pairs filtered by both metrics were 57\%, 62\%, 67\%, 71\%, and 76\%, respectively.  
We then randomly sampled 10 image–caption pairs that did not overlap between the two metrics’ filtered sets at the 30\% filtering rate, which was the setting that yielded the best performance when using the F-CLIPScore metric.  
The results are presented in Figure~\ref{fig:30_images}.
As shown in Figure~\ref{fig:30_images}, the images filtered by CLIPScore (upper row) appear to include some that should not have been filtered out, despite having normal and semantically correct captions. 
Although cases such as (g) and (h) involve repetitive lexical usage—where a word is repeated even if it correctly refers to an object present in the image—it is still reasonable for such captions to receive lower scores.  
However, it is unfortunate that these samples were filtered only by CLIPScore and not by F-CLIPScore.

On the other hand, the images filtered by F-CLIPScore (lower row in Figure~\ref{fig:30_images}) include cases such as (l), where even though there is repetitive lexical usage, the corresponding object in the image is actually a model rather than a designer, making the filtering reasonable.  
There were also examples like (n), where the captioned object is missing from the image, and (r), where the image shows a personal debt chart for “Brainy” rather than a “brain”.  
These qualitative samples provide insights into how filtering with F-CLIPScore influenced the performance of the LVLMs.

\begin{figure}[tbh]
  \centering

  \begin{minipage}{0.8\linewidth}
    \centering

    \begin{subfigure}[t]{0.17\linewidth}
      \centering
      \includegraphics[width=\linewidth]{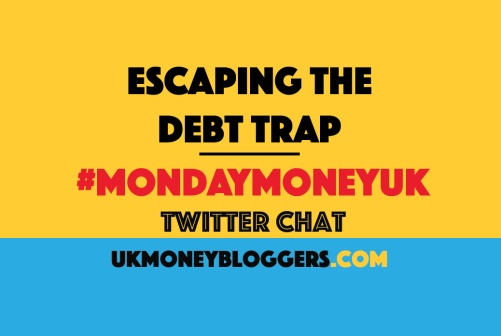}
      \caption{\scriptsize the image shows a photo of a money trap with text escaping the debt trap mondaysmoneyuk twitter chat}
    \end{subfigure}\hfill
    \begin{subfigure}[t]{0.17\linewidth}
      \centering
      \includegraphics[width=\linewidth]{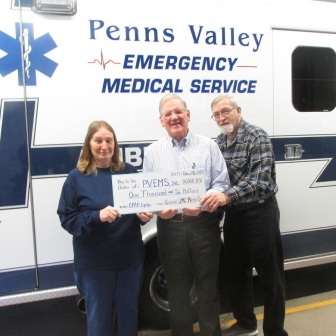}
      \caption{\scriptsize an elderly couple with a check presentation from the penns valley ambulance in michigan}
    \end{subfigure}\hfill
    \begin{subfigure}[t]{0.17\linewidth}
      \centering
      \includegraphics[width=\linewidth]{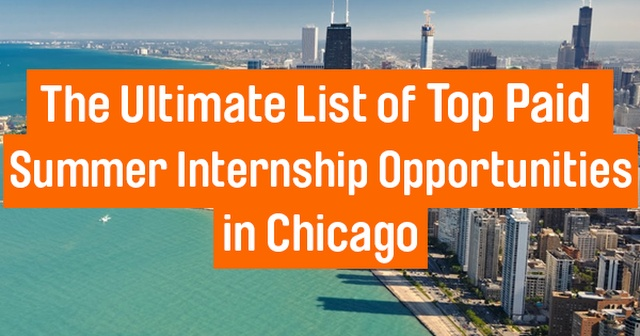}
      \caption{\scriptsize a lake michigan skyline and a chicago skyline with the text the ultimate list of top paid summer internship opportunities in}
    \end{subfigure}\hfill
    \begin{subfigure}[t]{0.17\linewidth}
      \centering
      \includegraphics[width=\linewidth]{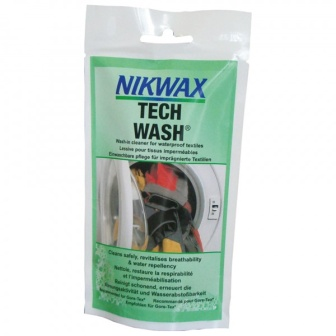}
      \caption{\scriptsize nikwax tech wash for bike helmets, helmets and helmets}
    \end{subfigure}\hfill
    \begin{subfigure}[t]{0.17\linewidth}
      \centering
      \includegraphics[width=\linewidth]{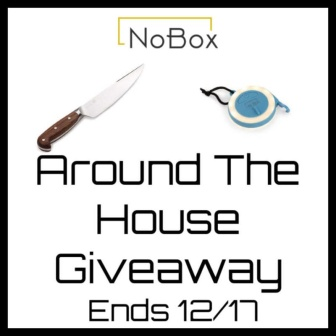}
      \caption{\scriptsize a knife, a clochet, and a round-up of the nobox household appliances gift give}
    \end{subfigure}

    \begin{subfigure}[t]{0.17\linewidth}
      \centering
      \includegraphics[width=\linewidth]{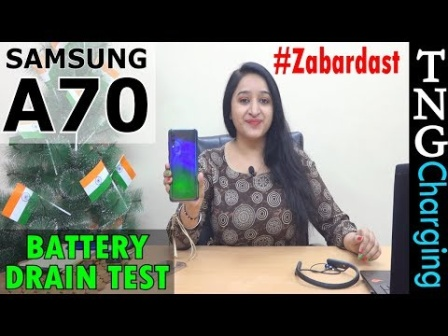}
      \caption{\scriptsize samsung a70 battery drain test, testing and test battery at home in 4 days | best battery indicator}
    \end{subfigure}\hfill
    \begin{subfigure}[t]{0.17\linewidth}
      \centering
      \includegraphics[width=\linewidth]{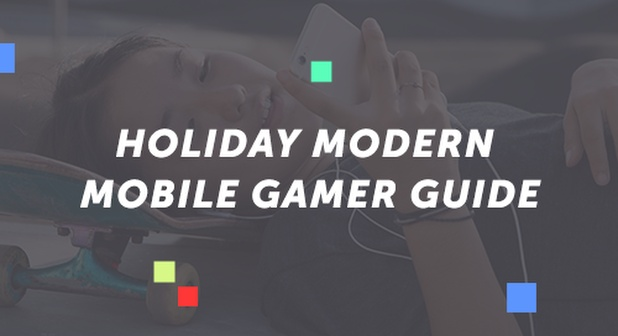}
      \caption{\scriptsize a girl lying on a skateboard with the words holiday modern mobile game guide in front}
    \end{subfigure}\hfill
    \begin{subfigure}[t]{0.17\linewidth}
      \centering
      \includegraphics[width=\linewidth]{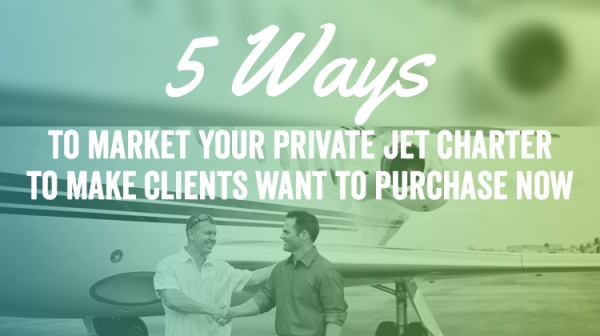}
      \caption{\scriptsize two men shaking hands outside of a plane with the title, 5 ways to market your private jet charter}
    \end{subfigure}\hfill
    \begin{subfigure}[t]{0.17\linewidth}
      \centering
      \includegraphics[width=\linewidth]{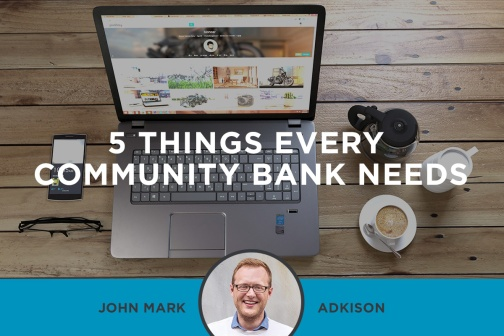}
      \caption{\scriptsize a laptop, phone, coffee cup, and keyboard with the words 5 things every community bank needs}
    \end{subfigure}\hfill
    \begin{subfigure}[t]{0.19\linewidth}
      \centering
      \includegraphics[width=\linewidth]{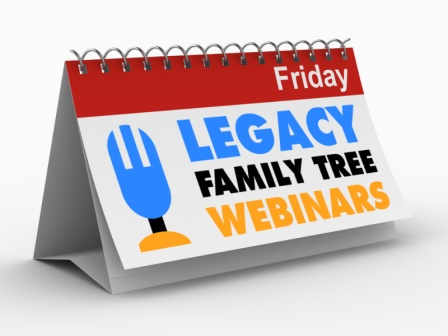}
      \caption{\scriptsize a calendar with the words legacy family tree webinars for a holiday meal}
    \end{subfigure}

    \vspace{0.5em}(I) Top 10 samples with the largest positive ranking differences\par\vspace{0.5em}
    
  \end{minipage}

  \begin{minipage}{0.8\linewidth}
    \centering

    \begin{subfigure}[t]{0.17\linewidth}
      \centering
      \includegraphics[width=\linewidth]{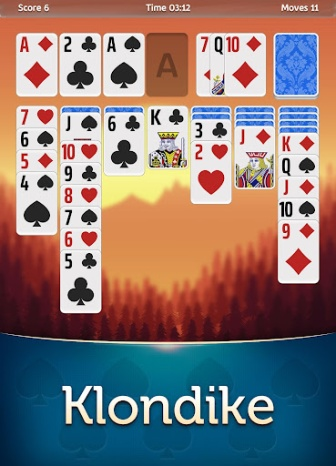}
      \caption{\scriptsize solitaire solita solitaire solita solitaire solitaire solitaire solitaire solitaire solitaire sol}
    \end{subfigure}\hfill
    \begin{subfigure}[t]{0.17\linewidth}
      \centering
      \includegraphics[width=\linewidth]{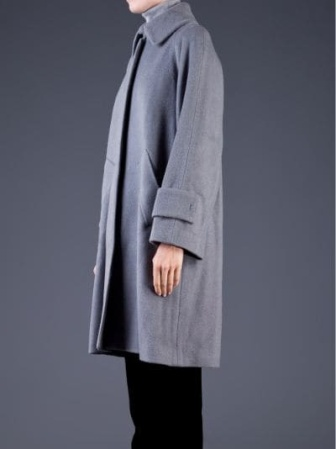}
      \caption{\scriptsize alexander alexander - double breasted coat 9}
    \end{subfigure}\hfill
    \begin{subfigure}[t]{0.17\linewidth}
      \centering
      \includegraphics[width=\linewidth]{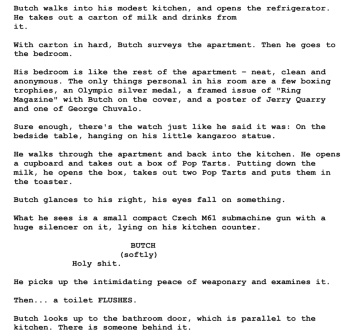}
      \caption{\scriptsize a screenplay screenplay with scriptwriting, handwritten}
    \end{subfigure}\hfill
    \begin{subfigure}[t]{0.17\linewidth}
      \centering
      \includegraphics[width=\linewidth]{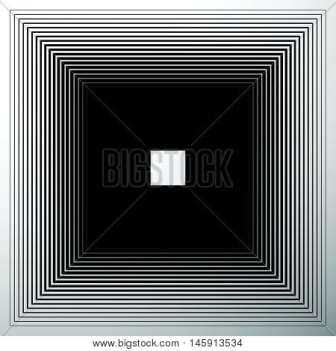}
      \caption{\scriptsize an illusion illusion illusion illusion in black and white}
    \end{subfigure}\hfill
    \begin{subfigure}[t]{0.17\linewidth}
      \centering
      \includegraphics[width=\linewidth]{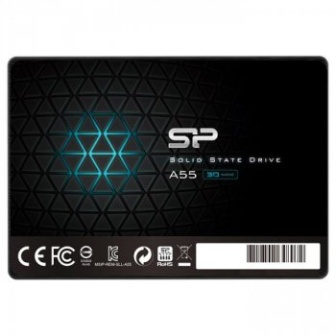}
      \caption{\scriptsize a52 250gb sat - 540 internal ssd}
    \end{subfigure}

    \begin{subfigure}[t]{0.17\linewidth}
      \centering
      \includegraphics[width=\linewidth]{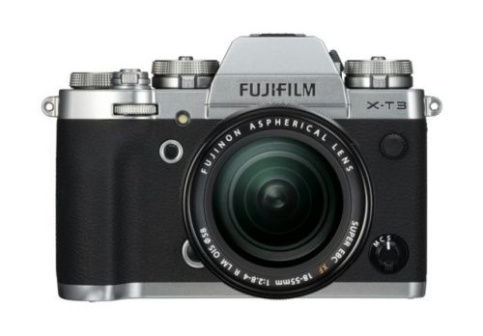}
      \caption{\scriptsize a fujifilm m7 with the fuji 105mm f2}
    \end{subfigure}\hfill
    \begin{subfigure}[t]{0.17\linewidth}
      \centering
      \includegraphics[width=\linewidth]{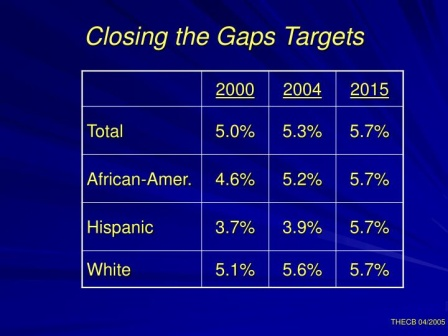}
      \caption{\scriptsize the gap and the gap of the gaps gap of the gaps gap on the gaps gap of the gaps is}
    \end{subfigure}\hfill
    \begin{subfigure}[t]{0.17\linewidth}
      \centering
      \includegraphics[width=\linewidth]{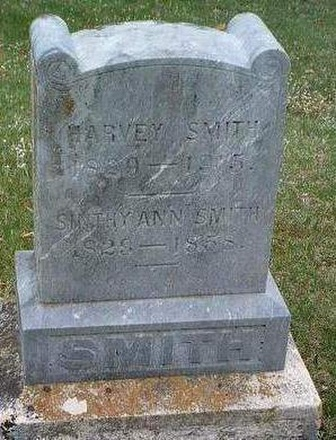}
      \caption{\scriptsize the headstone of robert s dutton}
    \end{subfigure}\hfill
    \begin{subfigure}[t]{0.17\linewidth}
      \centering
      \includegraphics[width=\linewidth]{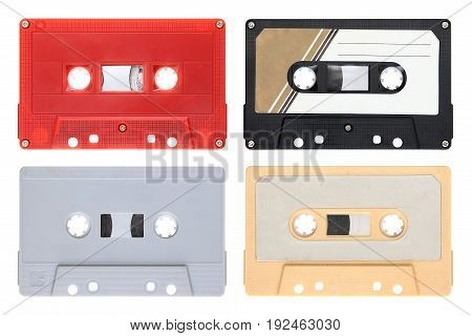}
      \caption{\scriptsize old cassette tape recorder}
    \end{subfigure}\hfill
    \begin{subfigure}[t]{0.17\linewidth}
      \centering
      \includegraphics[width=\linewidth]{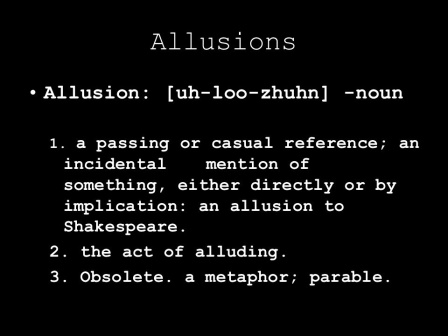}
      \caption{\scriptsize the algorithm, with an allusion}
    \end{subfigure}

    \vspace{0.5em}(II) Bottom 10 samples with the largest negative ranking differences\par

  \end{minipage}

  \caption{The figure shows the top and bottom 10 samples from the entire LLaVA-Pretrain dataset, sorted by the difference between the F-CLIPScore rank and the CLIPScore rank. (I) represents samples that CLIPScore rated higher than F-CLIPScore, while (II) represents the opposite cases. Each caption is written below its corresponding image.}
  \label{fig:whole_images}
\end{figure}

Furthermore, we measured the CLIPScore and F-CLIPScore for each image–caption pair across the entire LLaVA-Pretrain dataset and sorted them in ascending order, such that a higher score corresponds to a higher rank.  
We then sorted the samples by the difference between the two ranks (F-CLIPScore rank $-$ CLIPScore rank) and selected the top 10 samples with the largest positive values (i.e., those that CLIPScore rated higher than F-CLIPScore, shown in the upper row of Figure~\ref{fig:whole_images}) and the bottom 10 samples with the largest negative values (i.e., those that F-CLIPScore rated higher than CLIPScore, shown in the lower row of Figure~\ref{fig:whole_images}).

As shown in Figure~\ref{fig:whole_images}, the samples on side (I), which CLIPScore rated higher than F-CLIPScore, contain many text-rendered images.  
Conversely, the samples on side (II), which F-CLIPScore rated higher than CLIPScore, tend to exhibit duplicated lexical usage.  
These findings are consistent with the previous observations, and such a trade-off is expected to make determining an appropriate filtering rate more challenging.

\section{Licenses for the datasets and models used}

The LLaVA pre-training code is licensed under the Apache License, Version~2.0.
LLaMA 2 7B is released under the LLaMA Community License.
The LLaVA pretraining dataset is licensed under a combination of LAION, CC, and SBU licenses.
The OpenAI CLIP model is distributed under the MIT License.
COCO is available under the Creative Commons Attribution 4.0 License (CC BY 4.0).
Flickr30k is provided under a custom license that permits use for non-commercial research and/or educational purposes.
NoCaps is distributed under the Creative Commons Attribution 2.0 License (CC BY 2.0).
We used the datasets and models in accordance with their respective licenses.

\end{document}

%% file: math_commands.tex
\usepackage{amsmath,amsfonts,bm}

\def\eqref#1{equation~\ref{#1}}

\def\1{\bm{1}}

\DeclareMathAlphabet{\mathsfit}{\encodingdefault}{\sfdefault}{m}{sl}
\SetMathAlphabet{\mathsfit}{bold}{\encodingdefault}{\sfdefault}{bx}{n}



%% file: main.bib
@inproceedings{LLaVA,
 author = {Liu, Haotian and Li, Chunyuan and Wu, Qingyang and Lee, Yong Jae},
 booktitle = {Advances in Neural Information Processing Systems},
 editor = {A. Oh and T. Naumann and A. Globerson and K. Saenko and M. Hardt and S. Levine},
 pages = {34892--34916},
 publisher = {Curran Associates, Inc.},
 title = {Visual Instruction Tuning},
 url = {https://proceedings.neurips.cc/paper_files/paper/2023/file/6dcf277ea32ce3288914faf369fe6de0-Paper-Conference.pdf},
 volume = {36},
 year = {2023}
}

@misc{lvlm_hallucination_survey,
      title={A Survey on Hallucination in Large Vision-Language Models}, 
      author={Hanchao Liu and Wenyuan Xue and Yifei Chen and Dapeng Chen and Xiutian Zhao and Ke Wang and Liping Hou and Rongjun Li and Wei Peng},
      year={2024},
      eprint={2402.00253},
      archivePrefix={arXiv},
      primaryClass={cs.CV},
      url={https://arxiv.org/abs/2402.00253}, 
}

@inproceedings{eval_obj_hallucination_lvlm,
    title = "Evaluating Object Hallucination in Large Vision-Language Models",
    author = "Li, Yifan  and
      Du, Yifan  and
      Zhou, Kun  and
      Wang, Jinpeng  and
      Zhao, Xin  and
      Wen, Ji-Rong",
    editor = "Bouamor, Houda  and
      Pino, Juan  and
      Bali, Kalika",
    booktitle = "Proceedings of the 2023 Conference on Empirical Methods in Natural Language Processing",
    month = dec,
    year = "2023",
    address = "Singapore",
    publisher = "Association for Computational Linguistics",
    url = "https://aclanthology.org/2023.emnlp-main.20/",
    doi = "10.18653/v1/2023.emnlp-main.20",
    pages = "292--305"
}

@inproceedings{lcd_lvlm,
    title = "Mitigating Hallucinations in Large Vision-Language Models ({LVLM}s) via Language-Contrastive Decoding ({LCD})",
    author = "Manevich, Avshalom  and
      Tsarfaty, Reut",
    editor = "Ku, Lun-Wei  and
      Martins, Andre  and
      Srikumar, Vivek",
    booktitle = "Findings of the Association for Computational Linguistics: ACL 2024",
    month = aug,
    year = "2024",
    address = "Bangkok, Thailand",
    publisher = "Association for Computational Linguistics",
    url = "https://aclanthology.org/2024.findings-acl.359/",
    doi = "10.18653/v1/2024.findings-acl.359",
    pages = "6008--6022"
}

@inproceedings{ohd-caps,
    title = "Investigating and Mitigating Object Hallucinations in Pretrained Vision-Language ({CLIP}) Models",
    author = "Liu, Yufang  and
      Ji, Tao  and
      Sun, Changzhi  and
      Wu, Yuanbin  and
      Zhou, Aimin",
    editor = "Al-Onaizan, Yaser  and
      Bansal, Mohit  and
      Chen, Yun-Nung",
    booktitle = "Proceedings of the 2024 Conference on Empirical Methods in Natural Language Processing",
    month = nov,
    year = "2024",
    address = "Miami, Florida, USA",
    publisher = "Association for Computational Linguistics",
    url = "https://aclanthology.org/2024.emnlp-main.1016/",
    doi = "10.18653/v1/2024.emnlp-main.1016",
    pages = "18288--18301"
}

@inproceedings{icd_lvlm,
    title = "Mitigating Hallucinations in Large Vision-Language Models with Instruction Contrastive Decoding",
    author = "Wang, Xintong  and
      Pan, Jingheng  and
      Ding, Liang  and
      Biemann, Chris",
    editor = "Ku, Lun-Wei  and
      Martins, Andre  and
      Srikumar, Vivek",
    booktitle = "Findings of the Association for Computational Linguistics: ACL 2024",
    month = aug,
    year = "2024",
    address = "Bangkok, Thailand",
    publisher = "Association for Computational Linguistics",
    url = "https://aclanthology.org/2024.findings-acl.937/",
    doi = "10.18653/v1/2024.findings-acl.937",
    pages = "15840--15853"
}

@inproceedings{
robust_instruction_tuning,
title={Mitigating Hallucination in Large Multi-Modal Models via Robust Instruction Tuning},
author={Fuxiao Liu and Kevin Lin and Linjie Li and Jianfeng Wang and Yaser Yacoob and Lijuan Wang},
booktitle={The Twelfth International Conference on Learning Representations},
year={2024},
url={https://openreview.net/forum?id=J44HfH4JCg}
}

@inproceedings{
ciem,
title={{CIEM}: Contrastive Instruction Evaluation Method for Better Instruction Tuning},
author={Hongyu Hu and Jiyuan Zhang and Minyi Zhao and Zhenbang Sun},
booktitle={NeurIPS 2023 Workshop on Instruction Tuning and Instruction Following},
year={2023},
url={https://openreview.net/forum?id=HVduJbHSSO}
}

@article{spacy,
  author = {Matthew Honnibal and Ines Montani and Sofie Van Landeghem and Adriane Boyd},
  title = {spaCy: Industrial-strength Natural Language Processing in Python},
  year = {2020},
  doi = {10.5281/zenodo.1212303}
}

@inproceedings{clip,
  title={Learning transferable visual models from natural language supervision},
  author={Radford, Alec and Kim, Jong Wook and Hallacy, Chris and Ramesh, Aditya and Goh, Gabriel and Agarwal, Sandhini and Sastry, Girish and Askell, Amanda and Mishkin, Pamela and Clark, Jack and others},
  booktitle={International conference on machine learning},
  pages={8748--8763},
  year={2021},
  organization={PMLR}
}

@inproceedings{clipscore,
    title = "{CLIPS}core: A Reference-free Evaluation Metric for Image Captioning",
    author = "Hessel, Jack  and
      Holtzman, Ari  and
      Forbes, Maxwell  and
      Le Bras, Ronan  and
      Choi, Yejin",
    editor = "Moens, Marie-Francine  and
      Huang, Xuanjing  and
      Specia, Lucia  and
      Yih, Scott Wen-tau",
    booktitle = "Proceedings of the 2021 Conference on Empirical Methods in Natural Language Processing",
    month = nov,
    year = "2021",
    address = "Online and Punta Cana, Dominican Republic",
    publisher = "Association for Computational Linguistics",
    url = "https://aclanthology.org/2021.emnlp-main.595/",
    doi = "10.18653/v1/2021.emnlp-main.595",
    pages = "7514--7528"
}

@misc{laion,
      title={LAION-400M: Open Dataset of CLIP-Filtered 400 Million Image-Text Pairs}, 
      author={Christoph Schuhmann and Richard Vencu and Romain Beaumont and Robert Kaczmarczyk and Clayton Mullis and Aarush Katta and Theo Coombes and Jenia Jitsev and Aran Komatsuzaki},
      year={2021},
      eprint={2111.02114},
      archivePrefix={arXiv},
      primaryClass={cs.CV},
      url={https://arxiv.org/abs/2111.02114}, 
}

@inproceedings{datacomp,
title={DataComp: In search of the next generation of multimodal datasets},
author={Samir Yitzhak Gadre and Gabriel Ilharco and Alex Fang and Jonathan Hayase and Georgios Smyrnis and Thao Nguyen and Ryan Marten and Mitchell Wortsman and Dhruba Ghosh and Jieyu Zhang and Eyal Orgad and Rahim Entezari and Giannis Daras and Sarah M Pratt and Vivek Ramanujan and Yonatan Bitton and Kalyani Marathe and Stephen Mussmann and Richard Vencu and Mehdi Cherti and Ranjay Krishna and Pang Wei Koh and Olga Saukh and Alexander Ratner and Shuran Song and Hannaneh Hajishirzi and Ali Farhadi and Romain Beaumont and Sewoong Oh and Alex Dimakis and Jenia Jitsev and Yair Carmon and Vaishaal Shankar and Ludwig Schmidt},
booktitle={Thirty-seventh Conference on Neural Information Processing Systems Datasets and Benchmarks Track},
year={2023},
url={https://openreview.net/forum?id=dVaWCDMBof}
}

@misc{qwen2vl,
      title={Qwen2-VL: Enhancing Vision-Language Model's Perception of the World at Any Resolution}, 
      author={Peng Wang and Shuai Bai and Sinan Tan and Shijie Wang and Zhihao Fan and Jinze Bai and Keqin Chen and Xuejing Liu and Jialin Wang and Wenbin Ge and Yang Fan and Kai Dang and Mengfei Du and Xuancheng Ren and Rui Men and Dayiheng Liu and Chang Zhou and Jingren Zhou and Junyang Lin},
      year={2024},
      eprint={2409.12191},
      archivePrefix={arXiv},
      primaryClass={cs.CV},
      url={https://arxiv.org/abs/2409.12191}, 
}

@misc{minigpt4,
      title={MiniGPT-4: Enhancing Vision-Language Understanding with Advanced Large Language Models}, 
      author={Deyao Zhu and Jun Chen and Xiaoqian Shen and Xiang Li and Mohamed Elhoseiny},
      year={2023},
      eprint={2304.10592},
      archivePrefix={arXiv},
      primaryClass={cs.CV},
      url={https://arxiv.org/abs/2304.10592}, 
}

@INPROCEEDINGS{internvl,
  author={Chen, Zhe and Wu, Jiannan and Wang, Wenhai and Su, Weijie and Chen, Guo and Xing, Sen and Zhong, Muyan and Zhang, Qinglong and Zhu, Xizhou and Lu, Lewei and Li, Bin and Luo, Ping and Lu, Tong and Qiao, Yu and Dai, Jifeng},
  booktitle={2024 IEEE/CVF Conference on Computer Vision and Pattern Recognition (CVPR)}, 
  title={Intern VL: Scaling up Vision Foundation Models and Aligning for Generic Visual-Linguistic Tasks}, 
  year={2024},
  volume={},
  number={},
  pages={24185-24198},
  doi={10.1109/CVPR52733.2024.02283}}

@article{llm_hallucination,
author = {Ji, Ziwei and Lee, Nayeon and Frieske, Rita and Yu, Tiezheng and Su, Dan and Xu, Yan and Ishii, Etsuko and Bang, Ye Jin and Madotto, Andrea and Fung, Pascale},
title = {Survey of Hallucination in Natural Language Generation},
year = {2023},
issue_date = {December 2023},
publisher = {Association for Computing Machinery},
address = {New York, NY, USA},
volume = {55},
number = {12},
issn = {0360-0300},
url = {https://doi.org/10.1145/3571730},
doi = {10.1145/3571730},
journal = {ACM Comput. Surv.},
month = mar,
articleno = {248},
numpages = {38}
}

@misc{sirenssong,
      title={Siren's Song in the AI Ocean: A Survey on Hallucination in Large Language Models}, 
      author={Yue Zhang and Yafu Li and Leyang Cui and Deng Cai and Lemao Liu and Tingchen Fu and Xinting Huang and Enbo Zhao and Yu Zhang and Yulong Chen and Longyue Wang and Anh Tuan Luu and Wei Bi and Freda Shi and Shuming Shi},
      year={2023},
      eprint={2309.01219},
      archivePrefix={arXiv},
      primaryClass={cs.CL},
      url={https://arxiv.org/abs/2309.01219}, 
}

@inproceedings{pope,
    title = "Evaluating Object Hallucination in Large Vision-Language Models",
    author = "Li, Yifan  and
      Du, Yifan  and
      Zhou, Kun  and
      Wang, Jinpeng  and
      Zhao, Xin  and
      Wen, Ji-Rong",
    editor = "Bouamor, Houda  and
      Pino, Juan  and
      Bali, Kalika",
    booktitle = "Proceedings of the 2023 Conference on Empirical Methods in Natural Language Processing",
    month = dec,
    year = "2023",
    address = "Singapore",
    publisher = "Association for Computational Linguistics",
    url = "https://aclanthology.org/2023.emnlp-main.20/",
    doi = "10.18653/v1/2023.emnlp-main.20",
    pages = "292--305"
}

@misc{llama2,
      title={Llama 2: Open Foundation and Fine-Tuned Chat Models}, 
      author={Hugo Touvron and Louis Martin and Kevin Stone and Peter Albert and Amjad Almahairi and Yasmine Babaei and Nikolay Bashlykov and Soumya Batra and Prajjwal Bhargava and Shruti Bhosale and Dan Bikel and Lukas Blecher and Cristian Canton Ferrer and Moya Chen and Guillem Cucurull and David Esiobu and Jude Fernandes and Jeremy Fu and Wenyin Fu and Brian Fuller and Cynthia Gao and Vedanuj Goswami and Naman Goyal and Anthony Hartshorn and Saghar Hosseini and Rui Hou and Hakan Inan and Marcin Kardas and Viktor Kerkez and Madian Khabsa and Isabel Kloumann and Artem Korenev and Punit Singh Koura and Marie-Anne Lachaux and Thibaut Lavril and Jenya Lee and Diana Liskovich and Yinghai Lu and Yuning Mao and Xavier Martinet and Todor Mihaylov and Pushkar Mishra and Igor Molybog and Yixin Nie and Andrew Poulton and Jeremy Reizenstein and Rashi Rungta and Kalyan Saladi and Alan Schelten and Ruan Silva and Eric Michael Smith and Ranjan Subramanian and Xiaoqing Ellen Tan and Binh Tang and Ross Taylor and Adina Williams and Jian Xiang Kuan and Puxin Xu and Zheng Yan and Iliyan Zarov and Yuchen Zhang and Angela Fan and Melanie Kambadur and Sharan Narang and Aurelien Rodriguez and Robert Stojnic and Sergey Edunov and Thomas Scialom},
      year={2023},
      eprint={2307.09288},
      archivePrefix={arXiv},
      primaryClass={cs.CL},
      url={https://arxiv.org/abs/2307.09288}, 
}

@misc{hacl,
      title={Hallucination Augmented Contrastive Learning for Multimodal Large Language Model}, 
      author={Chaoya Jiang and Haiyang Xu and Mengfan Dong and Jiaxing Chen and Wei Ye and Ming Yan and Qinghao Ye and Ji Zhang and Fei Huang and Shikun Zhang},
      year={2024},
      eprint={2312.06968},
      archivePrefix={arXiv},
      primaryClass={cs.CV},
      url={https://arxiv.org/abs/2312.06968}, 
}

@misc{davies2015wikipedia,
  author       = {Mark Davies},
  title        = {The Wikipedia Corpus},
  year         = {2015},
  howpublished = {\url{https://www.english-corpora.org/wiki/}},
  note         = {Available online}
}

@InProceedings{Goyal_2024_CVPR,
    author    = {Goyal, Sachin and Maini, Pratyush and Lipton, Zachary C. and Raghunathan, Aditi and Kolter, J. Zico},
    title     = {Scaling Laws for Data Filtering-- Data Curation cannot be Compute Agnostic},
    booktitle = {Proceedings of the IEEE/CVF Conference on Computer Vision and Pattern Recognition (CVPR)},
    month     = {June},
    year      = {2024},
    pages     = {22702-22711}
}

@inproceedings{
negclip,
title={When and Why Vision-Language Models Behave like Bags-Of-Words, and What to Do About It?},
author={Mert Yuksekgonul and Federico Bianchi and Pratyusha Kalluri and Dan Jurafsky and James Zou},
booktitle={The Eleventh International Conference on Learning Representations },
year={2023},
url={https://openreview.net/forum?id=KRLUvxh8uaX}
}

@inproceedings{ceclip,
  title={Contrasting intra-modal and ranking cross-modal hard negatives to enhance visio-linguistic compositional understanding},
  author={Zhang, Le and Awal, Rabiul and Agrawal, Aishwarya},
  booktitle={Proceedings of the IEEE/CVF Conference on Computer Vision and Pattern Recognition},
  pages={13774--13784},
  year={2024}
}

@inproceedings{siglip,
  title={Sigmoid loss for language image pre-training},
  author={Zhai, Xiaohua and Mustafa, Basil and Kolesnikov, Alexander and Beyer, Lucas},
  booktitle={Proceedings of the IEEE/CVF international conference on computer vision},
  pages={11975--11986},
  year={2023}
}

@article{eva-clip,
  title={Eva-clip: Improved training techniques for clip at scale},
  author={Sun, Quan and Fang, Yuxin and Wu, Ledell and Wang, Xinlong and Cao, Yue},
  journal={arXiv preprint arXiv:2303.15389},
  year={2023}
}

@inproceedings{spacyparser,
    title = "An Improved Non-monotonic Transition System for Dependency Parsing",
    author = "Honnibal, Matthew  and
      Johnson, Mark",
    editor = "M{\`a}rquez, Llu{\'i}s  and
      Callison-Burch, Chris  and
      Su, Jian",
    booktitle = "Proceedings of the 2015 Conference on Empirical Methods in Natural Language Processing",
    month = sep,
    year = "2015",
    address = "Lisbon, Portugal",
    publisher = "Association for Computational Linguistics",
    url = "https://aclanthology.org/D15-1162/",
    doi = "10.18653/v1/D15-1162",
    pages = "1373--1378"
}
